\documentclass[]{interact}

\usepackage{amsmath}
\usepackage{amssymb}
\usepackage{graphicx}
\usepackage{epstopdf}
\usepackage[caption=false]{subfig}
\usepackage{booktabs}
\usepackage{multirow}
\usepackage{adjustbox}
\usepackage{url}

\usepackage{natbib}
\bibpunct[, ]{(}{)}{;}{a}{}{,}

\begin{document}

\articletype{RESEARCH ARTICLE}

\title{Robust vessel trajectory prediction under long missing gaps}

\author{
\name{Sanha Chang\textsuperscript{a} and Heeyoung Kim\textsuperscript{a}\thanks{CONTACT Heeyoung Kim. Email: heeyoungkim@kaist.ac.kr}}
\affil{\textsuperscript{a}Department of Industrial and Systems Engineering, Korea Advanced Institute of Science and Technology (KAIST), Daejeon 34141, South Korea}
}

\maketitle

\begin{abstract}
Automatic Identification System (AIS) streams often contain long consecutive missing gaps that break trajectory continuity and complicate vessel trajectory prediction. Existing missing-aware models leave these gaps unfilled, using techniques such as padding or attention masking, but their performance deteriorates as observations become sparse. In contrast, imputation-aware models fill missing gaps and incorporate imputation indicators, alleviating data scarcity but risking error propagation when imputed observations attend to one another across extended gaps. To address these challenges, this study proposes the Dynamic Imputation-Discounting Attention Mask Network (DIDAM-Net) for AIS trajectory prediction. DIDAM-Net learns a soft attention mask from missing-value patterns and contextual information to adaptively down-weight unreliable imputed observations, allowing them to provide auxiliary information without dominating prediction. 
By preserving trajectory continuity while limiting the influence of imputation errors, DIDAM-Net improves prediction accuracy under consecutive missing gaps. Experiments on real-world AIS data demonstrate that DIDAM-Net consistently reduces prediction errors and maintains robust performance under severe data sparsity and across different imputation schemes.
\end{abstract}

\begin{keywords}
Vessel trajectory prediction; Automatic Identification System; missing data; attention mechanism; autonomous shipping
\end{keywords}

\section{Introduction}
Accurate vessel trajectory prediction is essential for voyage optimization, autonomous surface vessel operations, and maritime traffic management.
With nearly 90\% of global commerce transported by sea, the scale of maritime trade further underscores the importance of reliable trajectory prediction \citep{harati2007automatic, yu2021literature}.
Vessel trajectory prediction primarily relies on Automatic Identification System (AIS) data, which are continuously broadcast by vessels worldwide. 
However, AIS streams often contain substantial vessel-specific missingness and long consecutive gaps caused by satellite or shore-station outages, human error, and environmental interference \citep{emmens2021promises}. 
In practical traffic management, operators must simultaneously monitor dozens to hundreds of vessels with heterogeneous reporting patterns, making robust prediction under incomplete observations particularly important  \citep{kaluza2010complex}.

Most existing vessel trajectory prediction studies handle missing AIS data by either filling gaps through simple imputation, typically linear or cubic interpolation, or restricting training to vessels with relatively complete AIS records \citep{donandt2023improved,wang2023adversarial, han2023interaction, wang2024vessel}. 
Although the latter produces cleaner datasets, it does not reflect operational settings in which vessels with diverse missing-data patterns must be monitored. 
The former, meanwhile, often treats imputed positions as genuine observations.
When actual vessel motion deviates from the interpolated trajectory, particularly across long consecutive gaps, imputation errors can distort the learned motion patterns and degrade prediction accuracy  \citep{bovsnjak2012automatic, han2023interaction}. 
These limitations motivate prediction methods that retain incomplete trajectories while explicitly accounting for the potentially unreliable information introduced by imputation.

Recent advances in neural trajectory prediction have motivated \textit{missing-aware} models that directly accommodate incomplete observation histories.
These models preserve missingness information through  padding tokens or attention masks, allowing the network to learn from sparse observations or perform internal imputation \citep{du2023saits, chib2024ms, liu2024mstf}. However, their prediction accuracy can deteriorate as consecutive missing gaps increase, because the remaining observations provide insufficient information to reliably infer vessel motion \citep{xu2023uncovering, liu2024mstf}.
This challenge is particularly relevant to AIS trajectories, where prolonged reporting gaps can leave only a few observations available for reconstructing recent motion and predicting future positions.

An alternative approach is \textit{imputation-aware} modeling, which has been explored in time-series forecasting but has received comparatively less attention in vessel trajectory prediction.
Unlike missing-aware models that retain unfilled gaps, imputation-aware models first fill missing timestamps and then provide the prediction model with information identifying imputed values.
This approach restores a complete temporal input sequence and can partially alleviate observation scarcity, but it also introduces the risk of propagating imputation errors  \citep{chib2024pedestrian}.
A common strategy augments imputed sequences with binary indicators distinguishing observed and imputed values.
However, such indicators do not capture differences in reliability among imputed points, particularly those within long consecutive gaps \citep{wang2024muse}.
More sophisticated approaches estimate uncertainty using Bayesian or ensemble-based techniques, but their additional computational demands can limit their suitability for  real-time trajectory prediction \citep{mulyadi2021uncertainty, tan2020data, li2025uncertainty}.

To address this challenge, this study proposes the \textit{Dynamic Imputation-Discounting Attention Mask Network} (DIDAM-Net), an imputation-aware framework for vessel trajectory prediction under long consecutive missing gaps. The central idea is to adaptively discount unreliable imputed observations during attention computation, allowing the model to benefit from imputed trajectories without over-relying on potentially erroneous values. Rather than using a fixed attention mask or treating all imputed points equally, DIDAM-Net dynamically generates attention masks based on the missingness structure and temporal context of each input trajectory. This design aims to alleviate observation scarcity while mitigating the propagation of imputation errors.

DIDAM-Net comprises three main components. 
First, a lightweight Multi-Layer Perceptron (MLP) generates the Dynamic Imputation-Discounting Attention Mask (DIDAM) from imputation metadata, including gap duration, local observation density, and relative temporal position.
The resulting pairwise mask is added to the pre-softmax attention logits, adaptively reducing attention involving potentially unreliable imputed points while preserving information from observed positions \citep{heo2018uncertainty, kobayashi2020attention}.
Second, a Transformer-based encoder incorporates DIDAM into multi-head self-attention to capture long-range trajectory dependencies while limiting the propagation of imputation errors, particularly through interactions among imputed points \citep{lim2021temporal}. 
Third, a Long Short-Term Memory (LSTM) decoder autoregressively predicts future positions, using gated memory to capture temporal dynamics \citep{koo2024deep}.
Masked cross-attention connects the decoder to the encoded history, with discounting derived from DIDAM reducing reliance on imputed observations while retaining their supplementary information.
Together, these components enable DIDAM-Net to learn from incomplete trajectories while adaptively controlling the influence of imputation.

The main contributions of this work are:
\begin{itemize}
   \item This study proposes DIDAM-Net, an imputation-aware vessel trajectory prediction framework designed to address both observation scarcity and imputation error propagation under long consecutive gaps in AIS data.
  \item This study introduces a dynamically generated attention mask that uses imputation metadata and temporal context to adaptively discount potentially unreliable imputed observations, limiting their influence in both encoder self-attention and decoder cross-attention.
  \item Experiments on real-world AIS data demonstrate that DIDAM-Net consistently outperforms competing trajectory prediction methods across different imputation strategies, with particularly strong performance under severe missingness and longer prediction horizons.
\end{itemize}

\section{Related work}
\subsection{Trajectory prediction}
Trajectory prediction has been widely studied across domains, including autonomous driving and pedestrian motion forecasting. 
Early Recurrent Neural Networks (RNNs), such as LSTM and Gated Recurrent Unit (GRU), captured temporal dependencies but struggled with long-range context and multi-agent interactions \citep{che2018recurrent,kim2023contextual}. 
Subsequent methods incorporated social interactions through social pooling (Social-LSTM), adversarial learning (Social-GAN), implicit social reasoning (Social-Implicit), and social-physical attention (SoPhie) \citep{alahi2016social, gupta2018social, mohamed2022social, sadeghian2019sophie}. 
Graph-based approaches, including STGAT, Social-STGCNN, and SGCN, further modeled agent interactions through spatio-temporal graph representations \citep{huang2019stgat, mohamed2020social, shi2021sgcn}. 
The Trajectron family combined probabilistic trajectory prediction with dynamic spatio-temporal graphs, with subsequent extensions in Trajectron++ and Smooth Trajectron++ \citep{ivanovic2019trajectron, salzmann2020trajectron++, westerhout2023smooth}.

More recently, Transformer-based architectures have enabled parallel sequence modeling and  long-range dependency learning through attention mechanisms  \citep{vaswani2017attention,koo2026gt}. Methods such as STAR, AgentFormer, and TUTR have advanced trajectory prediction by incorporating spatio-temporal context, agent interactions, and multimodal prediction \citep{yu2020spatio, yuan2021agentformer, shi2023trajectory}.
However, many existing methods assume complete or nearly complete observation histories, limiting their applicability to trajectories with long consecutive missing gaps. Recent studies have addressed prediction under incomplete  observations \citep{gu2022stochastic, xu2023uncovering, chib2024ms, liu2024mstf}, but their performance can still deteriorate as missing intervals increase.
Moreover, evaluations often rely on fully observed trajectories to construct missing-data scenarios and provide prediction targets. These limitations motivate trajectory prediction methods that can effectively exploit incomplete observation histories while mitigating the influence of potentially unreliable imputed values.

\subsection{Vessel trajectory prediction}
Vessel trajectory prediction relies primarily on AIS data  and faces challenges specific to maritime environments, including irregular sampling, missing observations, and navigation constraints imposed by Traffic Separation Schemes (TSS) and speed restrictions \citep{bovsnjak2012automatic, kim2017early, emmens2021promises}. 
Classical kinematic approaches, such as constant-velocity and constant-turn models and Kalman filters, remain effective for short-horizon prediction but struggle with  abrupt course changes and sparse observations  \citep{perera2010ocean}. 
Hybrid methods combine kinematic motion models with data-driven learning, including knowledge-assisted particle filtering with adaptive training-window selection  \citep{xiao2020big}.
More recent studies have adopted deep learning architectures, ranging from single-vessel sequence models to interaction-aware approaches incorporating Transformers and spatio-temporal attention mechanisms \citep{li2023ais, huang2024vessel, wang2024tptrans, li2024enhancing}. 
Route priors and shipping-lane information have also been explored to improve long-horizon prediction, while computational efficiency remains important for operational deployment \citep{kaluza2010complex, liang2024aisclean}.

Despite these advances, handling incomplete AIS observations remains a significant challenge. 
Many existing approaches either pre-impute missing positions or discard trajectories with extensive missing data  \citep{donandt2023improved, han2023interaction}. Pre-imputation can introduce substantial errors over long gaps, particularly when imputed positions are treated as genuine observations. More recent approaches combine probabilistic prediction, such as multi-output Gaussian processes, with historical-route information to improve robustness to missing observations and noise, although at increased computational cost \citep{wang2024ais}.
Nevertheless, relatively little attention has been given to explicitly controlling the influence of potentially unreliable imputed positions during trajectory prediction. This motivates DIDAM-Net, which adaptively discounts imputed observations based on missingness structure and temporal context while retaining their supplementary information. Its computational feasibility for real-time operation is evaluated in Section~\ref{sec:cost}.

\subsection{Attention masks in time-series prediction}
Attention masks are widely used in time-series models  to regulate information exchange across timestamps, particularly when observations are missing or noisy. Existing masking strategies differ in how they restrict interactions involving missing values. SAITS and MS-TIP employ diagonal self-masking to prevent timestamps from attending to themselves \citep{du2023saits, chib2024ms}, whereas more restrictive approaches exclude missing positions from attention queries or keys, preventing information exchange involving these positions \citep{chen2022combining, neog2025masking}.

For imputed sequences, attention masking can also account for the potential unreliability of imputed values. Some approaches apply diagonal masking to imputed positions  \citep{li2025uncertainty}, while others explicitly estimate imputation uncertainty and incorporate it into attention masks \citep{tan2020data}. However, diagonal masking does not prevent error propagation through interactions among imputed values, whereas explicit uncertainty estimation can introduce substantial computational overhead. 
Alternatively, DecayAttention \citep{zhang2024improved} generates temporal-distance-based decay weights and applies them to attention outputs after softmax. Although computationally efficient, this approach relies solely on temporal distance and does not account for the structure or context of missing gaps. 

To address these limitations, DIDAM uses a lightweight MLP to dynamically generate pairwise attention masks, including missing-block structure, local observation density, and temporal distance. Unlike temporal-distance-based decay, DIDAM incorporates multiple cues associated with imputation reliability and directly modulates pre-softmax attention scores without explicit uncertainty estimation. By adaptively down-weighting interactions involving imputed observations, particularly those within the same missing block, DIDAM retains useful contextual information while limiting error propagation across long consecutive gaps.

\section{Methodology}
\label{sec:methodology}

\subsection{Problem formulation}
Let $t \in \{1,\dots,T_{\mathrm{pred}}\}$ be the discrete time index, where $1{:}T_{\mathrm{obs}}$ denotes the observation history and $T_{\mathrm{obs}}{+}1{:}T_{\mathrm{pred}}$ denotes the prediction horizon, with $T_{\mathrm{obs}} < T_{\mathrm{pred}}$. 
For trajectory sequence $i$, the geographic position at time $t$ is denoted by
\begin{equation}
    \mathbf{y}_i^{t} = (\lambda_i^{t}, \phi_i^{t}) \in \mathbb{R}^2,
\end{equation}
where $\lambda_i^{t}$ and $\phi_i^{t}$ are longitude and latitude in degrees,
respectively.
The historical and the ground-truth future trajectories are denoted by 
\begin{equation}
    \{\mathbf{y}_i^{t}\}_{t=1}^{T_{\mathrm{obs}}}
    \quad \text{and} \quad
    \{\mathbf{y}_i^{t}\}_{t=T_{\mathrm{obs}}+1}^{T_{\mathrm{pred}}},
\end{equation}
respectively. The corresponding predicted future trajectory is
\begin{equation} \{\hat{\mathbf{y}}_i^{t}\}_{t=T_{\mathrm{obs}}+1}^{T_{\mathrm{pred}}}, 
\end{equation} 
where $\hat{\mathbf{y}}_i^{t} = (\hat{\lambda}_i^{t}, \hat{\phi}_i^{t})$ denotes the predicted position at time $t$.

\begin{figure}[htbp]
    \centering
    \includegraphics[width=0.45\textwidth]{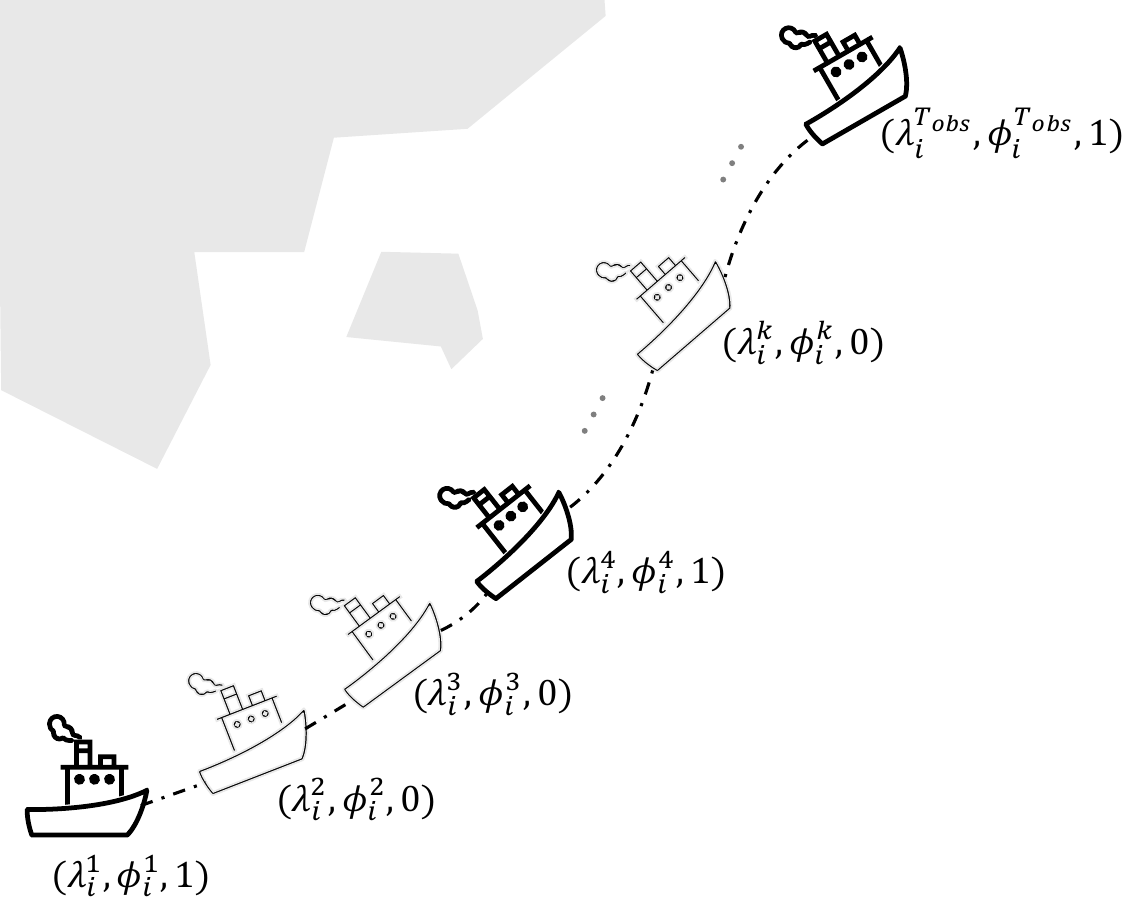}
    \caption{Example of a historical trajectory with observed and imputed positions and their corresponding imputation indicators $ m_i^t$.}
    \label{fig:input_example}
\end{figure}

Each historical trajectory is accompanied by a binary  \emph{imputation indicator}
\begin{equation}
\{m_i^{t}\}_{t=1}^{T_{\mathrm{pred}}}, \quad m_i^{t} \in \{0,1\},
\end{equation}
where $m_i^{t}=1$ indicates an observed AIS position and $m_i^{t}=0$ indicates an imputed position.
This indicator is generated during preprocessing and used to identify consecutive missing gaps and construct the imputation metadata for DIDAM.
Figure~\ref{fig:input_example} illustrates an example of a historical trajectory and its corresponding imputation indicators.

\subsection{Training objective}
To avoid supervising predictions with potentially inaccurate imputed values,  the prediction loss is minimized over observed future timestamps only:
\begin{equation}
\mathcal{L}
= \frac{1}{\sum_{i=1}^M \sum_{t=T_{\mathrm{obs}}+1}^{T_{\mathrm{pred}}} m_i^t}
  \sum_{i=1}^M \sum_{t=T_{\mathrm{obs}}+1}^{T_{\mathrm{pred}}} 
  m_i^t \,\|\hat{\mathbf{y}}_i^{t}-\mathbf{y}_i^{t}\|_2,
\end{equation}
where $M$ is the mini-batch size and $m_i^t=1$ if the future position at time $t$ is observed and $0$ otherwise.
By excluding imputed positions from supervision, the loss prevents the model from fitting potentially erroneous imputation targets and focuses learning on actual vessel observations.

\subsection{Model architecture}
\subsubsection{Overview of the architecture}
\label{sec:architecture_overview}

\begin{figure}[htbp]
\centering
\includegraphics[width=1.0\textwidth]{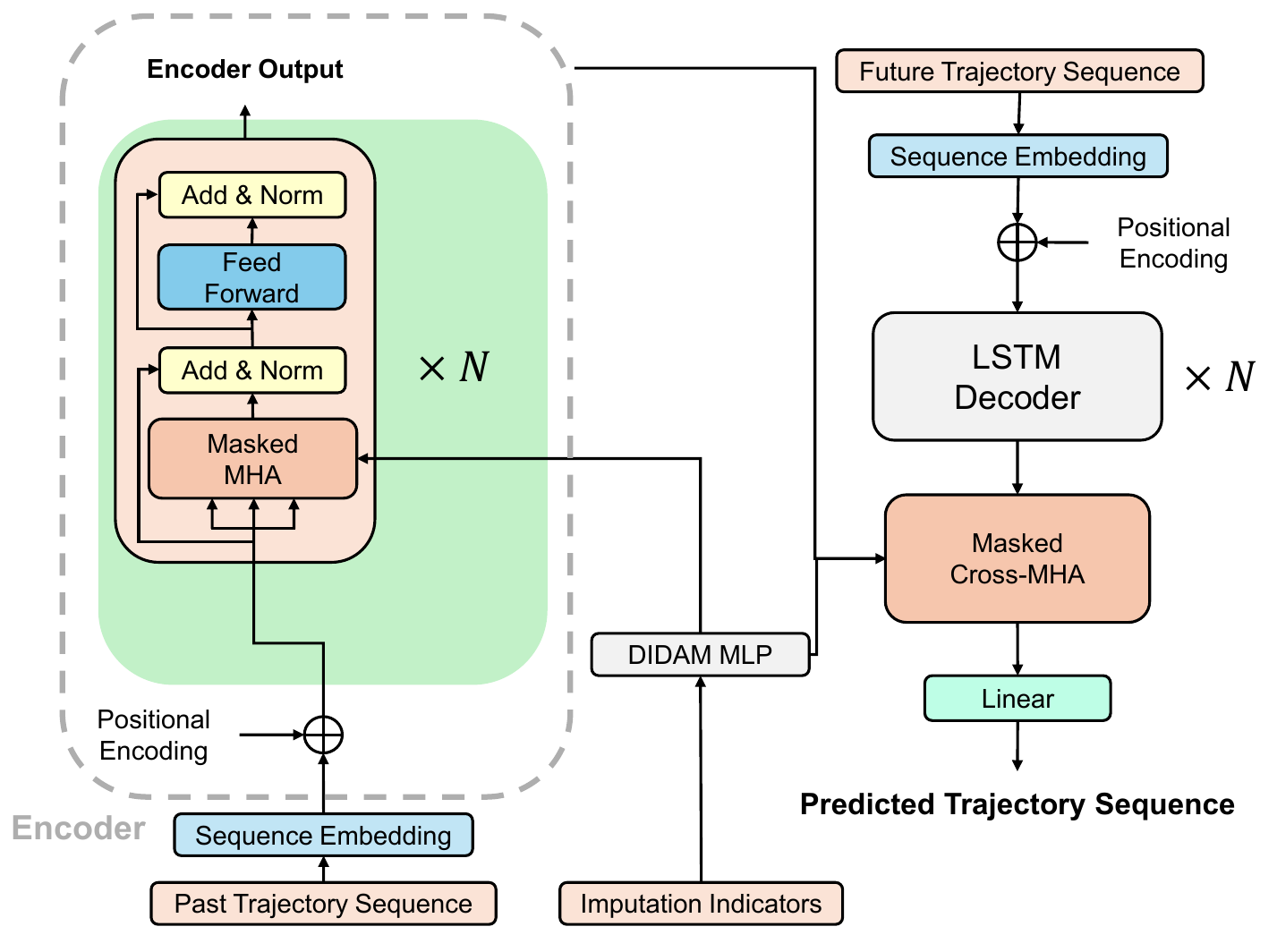}
\caption{Overview of DIDAM-Net.  The Transformer encoder (left) processes embedded historical trajectories through $N$ layers of masked self-attention. A lightweight MLP generates DIDAM from imputation metadata, which is incorporated into both encoder self-attention and decoder cross-attention. The LSTM decoder (right) uses the encoded history to predict future vessel positions.}
\label{fig:architecture}
\end{figure}

Figure~\ref{fig:architecture} illustrates the overall architecture of DIDAM-Net, which integrates a Transformer-based encoder and an LSTM decoder through the DIDAM mechanism.
DIDAM-Net takes as input historical trajectory coordinates  $(\lambda_i^t, \phi_i^t)$ and binary imputation indicators $m_i^t$, where 1 denotes an observed position or 0 an imputed position. The input coordinates are embedded and combined with positional encodings to retain temporal information.
DIDAM-Net consists of three main components: 
(1) a lightweight MLP that generates DIDAM from imputation metadata, 
(2) a Transformer encoder with masked self-attention, and (3) an LSTM decoder with masked cross-attention. 
DIDAM adaptively discounts potentially unreliable imputed positions in both attention mechanisms, allowing the model to exploit imputed trajectories while limiting error propagation. The following subsections describe each component in detail.

\subsubsection{DIDAM MLP}
The DIDAM MLP generates a pairwise attention discounting mask
$W \in (0,1]^{T_{\text{obs}} \times T_{\text{obs}}}$ from imputation metadata.
The mask is defined over temporal positions $t,t' \in \{1,\dots,T_{\text{obs}}\}$.
Each entry $W_{t,t'}$ determines the discount applied to attention from position $t$ (query) to position $t'$ (key).
DIDAM specifically discounts interactions between imputed positions within the same missing block, defined as a maximal consecutive sequence of imputed positions.

For input trajectory $i$, missing blocks are identified from the imputation indicators $\{m_i^t\}_{t=1}^{T_{\mathrm{obs}}}$. 
Let $\mathcal{B} = \{B^{(k)}\}_{k=1}^{N_{\mathrm{blk}}}$ denote the resulting $N_{\mathrm{blk}}$ missing blocks, where
\[
B^{(k)} = \{t_k, t_k{+}1, \ldots, t'_k\}, \quad
m_i^t = 0 \ \text{for all } t \in B^{(k)}.
\]
Here, $N_{\mathrm{blk}} = 0$ corresponds to a fully observed sequence, whereas larger $N_{\mathrm{blk}}$ values
indicate many shorter gaps scattered over time.

For each pair of temporal positions $(t,t')$, an imputation metadata vector $\boldsymbol{\psi}_{t,t'}$ is constructed from the missing-block structure and surrounding observations. The metadata comprise three feature groups: 
\begin{itemize}
    \item \textbf{Block structure:} the position of $t$ within its missing block and the block length;
    \item \textbf{Temporal distance:} the temporal separation between $t$ and $t'$, and the distances from $t$ to its nearest preceding and following observed positions;
    \item \textbf{Context:} the lengths of neighboring missing blocks, the number of observed positions between $t$ and $t'$, and the relative position of $t$ within the sequence.
\end{itemize}
These features provide complementary proxies for imputation reliability. 
\emph{Block structure} characterizes the length of a missing gap and the position of each imputed timestamp relative to its observed boundaries. Imputation errors may become more strongly correlated within long gaps, particularly at positions farther from observed boundaries, motivating the down-weighting of attention between imputed timestamps in the same missing block.
\emph{Temporal distance} captures the separation between timestamps and their distances to the nearest observed anchors, reflecting the potential deterioration of imputation reliability with increasing distance from observed data. 
\emph{Context} summarizes surrounding observations and missing gaps, including the number of observed timestamps between two positions and the lengths of neighboring gaps. These features allow attention discounts to vary across missing blocks of similar lengths according to their local observational context.

\begin{figure}[htbp]
\centering
\includegraphics[width=0.7\columnwidth]{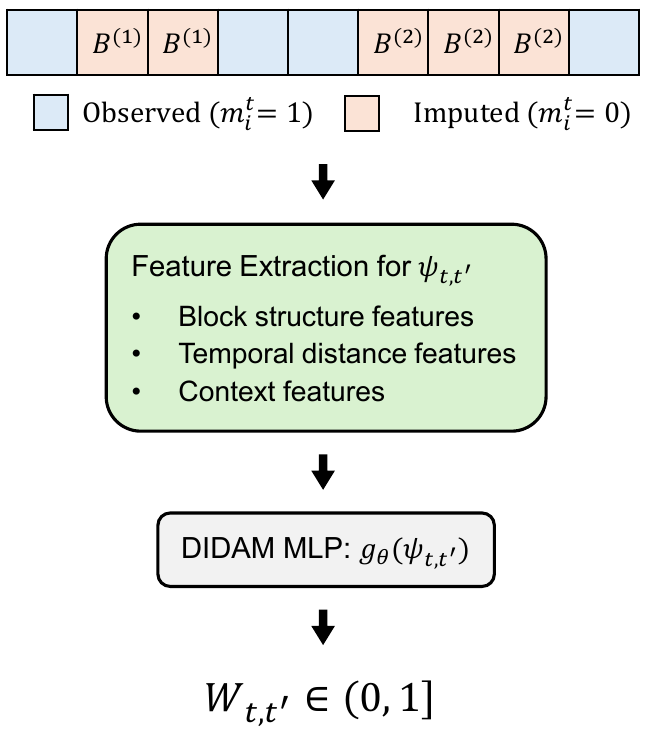}
\caption{Overview of DIDAM generation: an MLP maps imputation metadata to pairwise attention discounts $W_{t,t'}\in(0,1]$, applied only to \emph{imputed--imputed} pairs within the same missing block.}
\label{fig:didam}
\end{figure}

Finally, DIDAM maps the imputation metadata to the attention mask entries via an MLP:
\begin{equation}
W_{t,t'}=\begin{cases}
g_\theta(\boldsymbol{\psi}_{t,t'}), & m_i^{t}=m_i^{t'}=0, \ t,t'\in B^{(k)},\\[2pt]
1, & \text{otherwise,}
\end{cases}
\end{equation}
where $g_\theta$ is an MLP with a sigmoid output layer that maps the metadata to a discount factor in $(0,1]$.
The metadata $\boldsymbol{\psi}_{t,t'}$ depend only on the missingness pattern, including missing-block boundaries and distances to surrounding observations. The resulting mask $W$ therefore discounts attention using these patterns rather than the imputed coordinate values, allowing the same masking mechanism to be used with different imputation schemes.
Consequently, $W_{t,t'} < 1$ only for imputed--imputed pairs within the same missing block, while any entry involving at least one observed position remains at full weight ($W_{t,t'}=1$).
In other words, the resulting matrix $W$ compactly encodes which query--key pairs should be discounted in subsequent attention layers. 
Figure~\ref{fig:didam} illustrates the process of identifying missing blocks, extracting imputation metadata, and generating the attention mask.

\subsubsection{Encoder architecture}
The encoder consists of $N$ Transformer layers with multi-head self-attention and a position-wise feed-forward network. 
Given the discounting mask $W \in (0,1]^{T_{\mathrm{obs}}\times T_{\mathrm{obs}}}$ defined above, the encoder adds $\log W$ to the standard dot-product attention scores before applying softmax \citep{vaswani2017attention}. 

\begin{figure}[t]
  \centering
  \includegraphics[width=0.6\columnwidth]{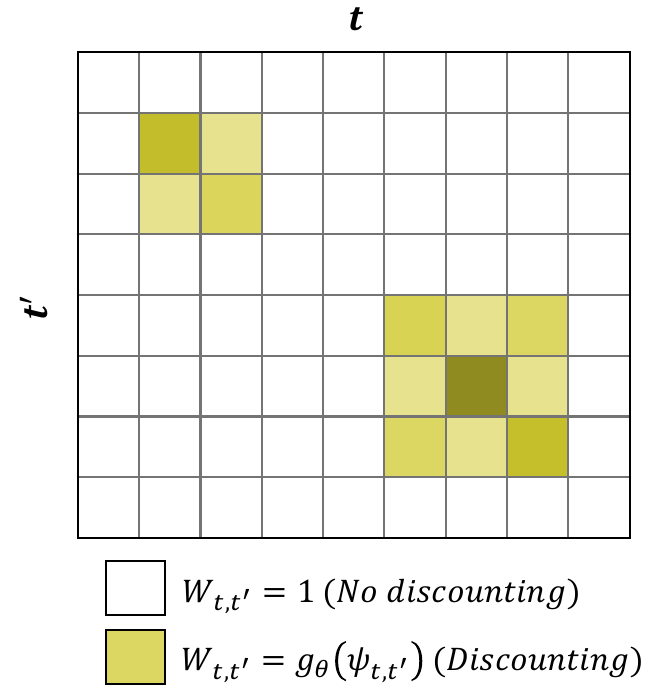}
  \caption{Encoder-side pairwise attention discounting mask $W$, which down-weights interactions between imputed positions within the same missing block.  }
  \label{fig:didam_W}
\end{figure}

Let $\tilde{\mathbf{A}} \in \mathbb{R}^{T_{\mathrm{obs}}\times T_{\mathrm{obs}}}$ denote
the DIDAM-modulated pre-softmax self-attention scores. Then, 
\begin{equation}
\tilde{\mathbf{A}} \;=\; \frac{\mathbf{Q}\mathbf{K}^{\top}}{\sqrt{d_k}} \;+\; \log W,
\end{equation}
where $\mathbf{Q}, \mathbf{K} \in \mathbb{R}^{T_{\mathrm{obs}}\times d_k}$ are the
query and key representations for a given head, and $d_k$ is the key dimensionality. 
The same mask $W$ is broadcast across all heads.
Adding $\log W$ before softmax is equivalent to multiplying the unnormalized attention weights by $W$ before row-wise normalization. 
As illustrated in Figure~\ref{fig:didam_W}, this \emph{soft masking} selectively discounts interactions between imputed positions within the same missing block. 
Pairs involving at least one observed position receive no direct discount ($W_{t,t'}=1$), although their normalized attention weights may change.  
Because $W_{t,t'}>0$, DIDAM reduces rather than eliminates attention between imputed positions, allowing the encoder to limit error propagation while retaining contextual information across the sequence.

\subsubsection{Decoder architecture with cross-attention}
Future positions are predicted autoregressively using stacked LSTM layers followed by cross-attention over the encoder outputs. The LSTM captures sequential motion dynamics, while cross-attention allows the decoder to incorporate information from the encoded history at each prediction step.
To incorporate DIDAM into cross-attention, a source discounting mask 
$R \in (0,1]^{(T_{\mathrm{pred}}-T_{\mathrm{obs}})\times T_{\mathrm{obs}}}$ is constructed by repeating the diagonal of the encoder-side mask $W$ across the prediction horizon: 
\[
R \;=\; \mathrm{row\mbox{-}broadcast}_{T_{\mathrm{pred}}-T_{\mathrm{obs}}}\bigl(\mathrm{diag}(W)\bigr).
\]
Each diagonal entry of $W$ provides a discount factor for the corresponding encoder position.  Thus, each encoder position receives the same discount across prediction steps, but different encoder positions can receive different discounts.

\begin{figure}[t]
  \centering
  \includegraphics[width=0.6\columnwidth]{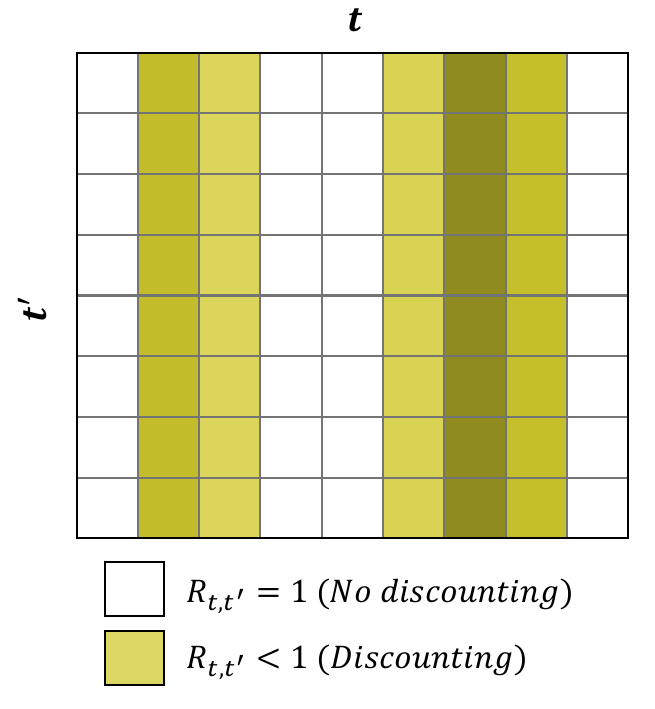}
  \caption{Decoder-side source discounting mask $R$, constructed by row-broadcasting $\mathrm{diag}(W)$ across the prediction horizon. Each row represents a decoder prediction step, and each column corresponds to an encoder position.}
  \label{fig:didam_R}
\end{figure}

Let $\tilde{\mathbf{A}}_{\times} \in \mathbb{R}^{(T_{\mathrm{pred}}-T_{\mathrm{obs}})\times T_{\mathrm{obs}}}$ 
denote the DIDAM-modulated pre-softmax cross-attention scores.
The log-mask is added to the pre-softmax cross-attention scores for all heads:
\begin{equation}
\tilde{\mathbf{A}}_{\times}
\;=\;
\frac{\mathbf{Q}_{\times}\mathbf{K}_{\times}^{\top}}{\sqrt{d_k}}
\;+\; \log R,
\end{equation}
where $\mathbf{Q}_{\times}$ and $\mathbf{K}_{\times}$ are the query and key matrices obtained from the decoder states and encoder outputs, respectively.
As in encoder self-attention, the log-mask multiplicatively discounts unnormalized attention weights before softmax normalization. The resulting mask reduces attention to encoder positions corresponding to imputed observations, while positions with observed AIS reports receive no direct discount. This allows the decoder to retain information from imputed positions while placing relatively greater emphasis on observed history. 
Figure~\ref{fig:didam_R} illustrates the decoder-side source discounting mask.

\section{Experiments}
\subsection{Datasets}
This study uses the Nationwide Automatic Identification System (NAIS) 2023 dataset provided by the National Oceanic and Atmospheric Administration Office for Coastal Management \citep{noaa_nais_2023}, collected on March 6--8, 2023, spanning latitudes $[29.50^\circ, 30.00^\circ]$ and longitudes $[-90.50^\circ, -89.50^\circ]$, with $155{,}457$ records from $293$ vessels.
The study area covers the Lower Mississippi River Delta near New Orleans, a major maritime gateway where dense vessel traffic and mixed coastal--offshore communication infrastructure often lead to frequent AIS missing gaps.
Preprocessing begins by removing records with abnormal speeds (Speed Over Ground, SOG, $\ge 30$ kn) after decoding the AIS `speed not available' code (102.3~kn) as a missing SOG value \citep{itu2014m1371}.
This threshold is well above typical commercial vessel speeds  \citep{nguyen2024transformer, paulig2024open}. No valid SOG report in the study area exceeds 20.3~kn (Table~\ref{tab:population}), so the threshold removes no records. A separate check of speeds implied by consecutive reports from the same vessel \citep{emmens2021promises, li2024enhancing} also identifies no abnormal case.
The preprocessing also removes reports from vessels that are not under way (SOG $\le 1$~kn), reports with COG or Heading marked as `not available,' and vessels that never report their length or width.
The remaining records are resampled onto a 1-minute grid by averaging multiple observations within each minute.

Each vessel's trajectory is then divided into sessions whenever a consecutive missing gap exceeds one hour, and sequences shorter than one hour are discarded. Each timestamp is assigned an imputation indicator $m_i^t$, where $m_i^t{=}1$ for originally observed positions and $m_i^t{=}0$ for missing positions subsequently imputed. 
Overall, $32.97\%$ of the trajectory timestamps require imputation, indicating substantial missingness in the AIS data. 

To prevent temporal data leakage, the resulting sessions are split chronologically into training, validation, and test sets in a 6:2:2 ratio. 
Missing positions are then imputed, and trajectory samples are extracted using a 60-timestamp (1-hour) sliding window with stride 1, consisting of 30 historical and 30 future timestamps. 
Samples with no originally observed positions in the prediction horizon are excluded from both training and evaluation, yielding $259{,}095$ trajectory samples.

Table~\ref{tab:population} summarizes the vessel population, which is dominated by large commercial or towing vessels operating at relatively low speeds. Vessel type and dimensions are not used as model inputs.

\begin{table}[t]
\centering
\tbl{Vessel population of the dataset by AIS ship-type group. Length and width are group medians of each vessel's most frequently reported values; mean SOG is the median of vessel-level mean speeds computed from retained moving reports; max SOG is the maximum single report within each group.}{%
\setlength{\tabcolsep}{4pt}
\begin{adjustbox}{max width=\textwidth}
\begin{tabular}{lrrrrr}
\toprule
Ship-type group (AIS codes) & Vessels & Length (m) & Width (m) & Mean SOG (kn) & Max SOG (kn) \\
\midrule
Tug / towing (31, 32, 52)        & 160 & 24  & 9  & 5.4  & 16.8 \\
Cargo (70--79)                   & 61  & 190 & 32 & 10.5 & 20.3 \\
Tanker (80--89)                  & 36  & 183 & 32 & 9.8  & 17.6 \\
Local vessel (56, 57)            & 21  & 122 & 18 & 4.5  & 14.1 \\
Passenger (60--69)               & 6   & 80  & 17 & 8.3  & 20.2 \\
Pleasure / sailing (36, 37)      & 2   & 24  & 7  & 9.2  & 14.7 \\
Other / not reported (0, 55, 90) & 7   & 43  & 10 & 4.8  & 12.6 \\
\midrule
Total                            & 293 & 37  & 12 & 6.3  & 20.3 \\
\bottomrule
\end{tabular}
\end{adjustbox}
}
\label{tab:population}
\end{table}

\subsection{Baseline methods}
DIDAM-Net is compared with representative trajectory prediction methods spanning statistical models, recurrent neural networks (RNNs), graph neural networks (GNNs), generative adversarial networks (GANs), and Transformer-based architectures. These baselines cover both established approaches, such as LSTM-based temporal models and spatio-temporal graph networks, and more recent Transformer-based methods.
Several baselines explicitly model interactions among neighboring vessels, providing a comparison with approaches that leverage additional inter-vessel context. This broad evaluation helps assess the effectiveness of DIDAM-Net's imputation-aware attention mechanism relative to different modeling paradigms.

\paragraph*{Extended Kalman filter (EKF).}
An EKF with a curvilinear motion model \citep{perera2010ocean} is used as a baseline, tuned on the validation set. 
The state includes position, velocity, and tangential/normal accelerations in geographic coordinates.
During missing gaps, the EKF performs prediction-only updates and resumes measurement updates when AIS observations become available. 
Although computationally efficient and effective for short-horizon prediction, the EKF may accumulate errors during complex maneuvers or prolonged observation gaps.

\paragraph*{Spatio-temporal graph attention network (STGAT).}
STGAT models inter-agent interactions through a dynamic graph with learned attention weights  \citep{huang2019stgat}. 
Multi-head graph attention layers adaptively aggregate information from neighboring agents, while a temporal encoder captures motion history for trajectory prediction. 
By selectively weighting neighboring agents, STGAT can exploit inter-vessel interactions in congested maritime environments.

\paragraph*{Social spatio-temporal graph convolution neural network (S-STGCNN).} 
S-STGCNN combines spatial graph convolutions and temporal convolutions to model multi-agent interactions and motion dynamics \citep{mohamed2020social}. 
A proximity-based graph represents interactions among neighboring agents, while graph convolutions aggregate spatial information and temporal convolutions capture motion patterns. Its convolutional architecture enables efficient trajectory prediction in densely populated environments.

\paragraph*{AgentFormer.}
AgentFormer is a multi-agent Transformer that jointly models agent trajectories using agent-aware attention to capture both intra- and inter-agent dependencies \citep{yuan2021agentformer}. Within a conditional variational autoencoder (CVAE) framework \citep{kingma2013auto}, it learns latent representations of future motion and autoregressively generates multimodal, socially consistent trajectories.

\paragraph*{Two-block RNN for incomplete trajectories (Two-block RNN).}
Two-block RNN extends a Social-GAN-based architecture \citep{gupta2018social} to handle incomplete trajectory observations. 
Its encoder alternates between two recurrent operations \citep{fujii2021two}: an update--prediction step when an observation is available and prediction-only propagation when it is missing. By processing missing observations without explicit imputation, the model avoids introducing imputation errors into its recurrent state updates.

\paragraph*{Transformer for trajectory forecasting (Transformer).}
This baseline employs a standard encoder--decoder Transformer to capture long-range temporal dependencies and predict future trajectories \citep{giuliari2021transformer}. 
Positional encodings preserve temporal order, while the decoder conditions on the encoded history to produce multi-step forecasts. 
Unlike interaction-aware models, this baseline does not explicitly model dependencies among neighboring agents.

\paragraph*{Social-Implicit (S-Implicit).}
S-Implicit models multi-agent interactions through implicit representations rather than explicit graph operations \citep{mohamed2022social}. 
It aggregates information from neighboring agents to construct interaction-aware representations for trajectory prediction, providing an alternative to graph-based social modeling.

\paragraph*{Multiscale Transformer for incomplete trajectories (MSTF).}
MSTF is a missing-aware Transformer designed for trajectory prediction with missing observations \citep{liu2024mstf}.
It replaces missing positions with zero padding and employs attention masks with different structural patterns to control information flow. By restricting attention involving missing positions, MSTF learns temporal dependencies primarily from available observations without explicitly imputing missing trajectories.

\paragraph*{Trajectory Unified Transformer (TUTR).}
TUTR is an encoder--decoder Transformer that integrates motion-mode reasoning and multi-agent interaction modeling \citep{shi2023trajectory}. It represents distinct future motion patterns through motion-mode tokens and jointly performs trajectory regression and mode classification. Its non-autoregressive decoder efficiently generates multiple plausible future trajectories.

\subsection{Implementation details}
\label{sec:implementation-details}
\paragraph*{Metrics.}
Model performance is evaluated using a best-of-$K$ protocol with $K=20$, a common evaluation strategy in trajectory prediction that accounts for multiple plausible future vessel trajectories.
At test time, for each trajectory sequence $i$, the model generates $K$ stochastic forecasted trajectories,
\[
\{\hat{\mathbf{y}}_i^{t,(k)}\}_{t=T_{\mathrm{obs}}+1}^{T_{\mathrm{pred}}}, \quad k \in \{1,\ldots,K\},
\]
for example via latent noise injection or stochastic dropout. Here, $k$ indexes stochastic trajectory samples, so that $\hat{\mathbf{y}}_i^{t,(k)}$ denotes the $k$-th forecasted trajectory for sequence $i$.

The Average Displacement Error (ADE) and Final Displacement Error (FDE), which are standard evaluation metrics in trajectory prediction, are adapted to handle missing future observations by leveraging the imputation indicator $m_i^t$.
The metrics select the best-performing sample among the $K$ predictions for each trajectory and then average across all trajectories:
\begin{equation}
    \mathrm{ADE} = \frac{1}{N_{\text{test}}}  \sum_{i=1}^{N_{\text{test}}}  \min_{k \in \{1, \ldots, K\}} \mathrm{ADE}_i^{(k)},
\end{equation}
\begin{equation}
    \mathrm{FDE} = \frac{1}{N_{\text{test}}}  \sum_{i=1}^{N_{\text{test}}}  \min_{k \in \{1, \ldots, K\}} \mathrm{FDE}_i^{(k)},
\end{equation}
where $N_{\text{test}}$ is the total number of test trajectories.
For the $k$-th prediction of trajectory $i$, $\mathrm{ADE}_i^{(k)}$ averages the great-circle distance over the \emph{observed} future timestamps, whereas $\mathrm{FDE}_i^{(k)}$ measures this distance at the \emph{last observed} future timestamp. They are defined as
\begin{equation}
    \mathrm{ADE}_i^{(k)} = \frac{1}{\sum_{t=T_{\mathrm{obs}}+1}^{T_{\mathrm{pred}}}m_i^t}\sum_{t=T_{\mathrm{obs}}+1}^{T_{\mathrm{pred}}} m_i^t \, d\bigl(\hat{\mathbf{y}}_i^{t,(k)}, \mathbf{y}_i^{t}\bigr),
\end{equation}
and
\begin{equation}
    \mathrm{FDE}_i^{(k)} = d\bigl(\hat{\mathbf{y}}_i^{t_\ast^{(i)},(k)}, \mathbf{y}_i^{t_\ast^{(i)}}\bigr),
\end{equation}
where 
 $   t_\ast^{(i)} = \max \{t \in \{T_{\mathrm{obs}}+1,\dots,T_{\mathrm{pred}}\} \mid m_i^t = 1\}
 $
is the index of the last future timestamp at which trajectory $i$
is actually observed. Accordingly, trajectory samples with no observed
timestamps in the prediction horizon were excluded from both training
and evaluation.
The distance $d$ between a predicted position $\hat{\mathbf{y}} = (\hat{\lambda}, \hat{\phi})$ and the corresponding true position $\mathbf{y} = (\lambda, \phi)$ is measured using the haversine great-circle distance:
\begin{equation}
    d(\hat{\mathbf{y}}, \mathbf{y}) = 2R \arcsin \sqrt{\sin^2\!\Bigl(\frac{\hat{\phi}-\phi}{2}\Bigr) + \cos\phi \, \cos\hat{\phi} \, \sin^2\!\Bigl(\frac{\hat{\lambda}-\lambda}{2}\Bigr)},
\end{equation}
where longitudes and latitudes are expressed in radians and $R = 6{,}371.0088$~km is the mean Earth radius. Distances are reported in nautical miles (NM; 1~NM $= 1{,}852$~m).
Unlike Euclidean distance computed directly from  coordinates in degrees, the haversine distance accounts for the latitude-dependent scaling of longitude: at the latitude of the study area, one degree of longitude corresponds to approximately 52~NM, compared with approximately 60~NM for one degree of latitude.

These metrics are used consistently across all experiments to evaluate performance strictly on \emph{observed} future points, ensuring that results reflect real-world predictive accuracy rather than imputation quality.
All reported performances are the means with standard errors computed over five random seeds.

\paragraph*{Training details.}
DIDAM-Net is implemented as described in Section~\ref{sec:methodology}, using 30 observation timestamps to predict the subsequent 30 timestamps.
Specifically, $T_{\mathrm{obs}}$ and $T_{\mathrm{pred}}$ are set to 30 and 60, respectively, so that the prediction horizon spans timestamps from $T_{\mathrm{obs}}+1$ to $T_{\mathrm{pred}}$.
Input coordinates are z-normalized using dataset-wise statistics (mean/variance), computed exclusively on the training split and then applied to the validation and test sets. 
The model predicts 2D displacements, which are cumulatively summed and de-normalized to recover absolute positions.
The DIDAM MLP is implemented as a two-layer network with a hidden dimension of 16.
The architecture employs a 512-dimensional embedding, 3 Transformer encoder layers, 3 LSTM decoder layers, 8-head attention, and a dropout rate of 0.1. 
The model is optimized with Adam using a Noam learning rate schedule initialized with embedding dimension 512, scaling factor 1.0, and warmup steps equal to 10 times the number of training batches per epoch, with $\beta_1 = 0.9$, $\beta_2 = 0.98$, and $\varepsilon = 10^{-9}$.  Training uses a batch size of 100, a maximum of 100 epochs, and early stopping (patience 30) based on validation loss computed over observed timestamps. 
The model parameters achieving the lowest validation loss are saved and used for the final evaluation on the test set. All experiments were conducted on a single NVIDIA TITAN V GPU (12\,GB) using PyTorch with CUDA 11.1 and Python 3.8.

\subsection{Results}
\label{sec:results}
Results are reported under two complementary evaluation settings based on the same raw AIS data but differing in the distribution of consecutive missing gaps. 
Unless otherwise stated, both settings follow the preprocessing procedures described above, including resampling, trajectory segmentation, filtering, normalization,
and chronological train/validation/test splitting.

The first experiment (\emph{Trajectory prediction under original missing gaps}, Table~\ref{tab:imputation_main}) evaluates the sensitivity of trajectory prediction models to different imputation strategies under the \emph{original} AIS missingness patterns. Each imputation method is applied to the preprocessed data without altering the missing blocks, and prediction performance is compared across models and imputation methods using ADE and FDE.

The second experiment (\emph{Trajectory prediction under prolonged missing gaps}, Table~\ref{tab:mingap5_allbaselines}) evaluates robustness to longer consecutive missing gaps.  
Starting from the same raw AIS data, missing blocks shorter than five timestamps are extended to at least five timestamps on the resampled time grid. The resulting dataset is then processed using the same segmentation, filtering, normalization, and chronological splitting procedures to construct a separate train/validation/test split. Five timestamps correspond to 5~min, exceeding the nominal AIS reporting interval of at most 3~min for vessels under way  \citep{itu2014m1371}. Existing gaps are extended in place rather than relocated, preserving their temporal locations. 
This duration also corresponds to the 90th percentile of the observed gap lengths, whose median is 1~min.  
The setting is designed as a robustness stress test rather than a simulation of real-world AIS outage statistics.

\providecommand{\best}[1]{\textbf{#1}}
\providecommand{\second}[1]{\underline{#1}}

\paragraph*{Trajectory prediction under original missing gaps.}
Table~\ref{tab:imputation_main} compares  ADE and FDE across several imputation strategies commonly used in vessel trajectory prediction \citep{donandt2023improved, han2023interaction, wang2023adversarial}. All methods are evaluated on the preprocessed AIS dataset with its original missingness patterns preserved. 
DIDAM-Net achieves the lowest prediction errors across all imputation strategies, with relatively small variations in performance. This consistency suggests that dynamic attention discounting reduces sensitivity to the choice of imputation method.

\begin{table}[t]
\centering
\tbl{Trajectory prediction performance under original missing gaps (mean $\pm$ SE of ADE$\downarrow$ / FDE$\downarrow$, in NM).}{%
\setlength{\tabcolsep}{2pt}
\renewcommand{\arraystretch}{1.2}
\begin{adjustbox}{max width=\textwidth}

\begin{tabular}{lcccc}
\toprule
Method & Linear & Cubic & Kalman filter & Gaussian process \\
\midrule
EKF 
& 3.124{\scriptsize $\pm$0.000} / 6.144{\scriptsize $\pm$0.000}
& 3.221{\scriptsize $\pm$0.000} / 6.345{\scriptsize $\pm$0.000}
& 2.669{\scriptsize $\pm$0.000} / 4.784{\scriptsize $\pm$0.000}
& 3.316{\scriptsize $\pm$0.000} / 6.423{\scriptsize $\pm$0.000} \\
STGAT 
& 1.362{\scriptsize $\pm$0.055} / 1.717{\scriptsize $\pm$0.061}
& 1.360{\scriptsize $\pm$0.047} / 1.714{\scriptsize $\pm$0.056}
& 1.375{\scriptsize $\pm$0.061} / 1.731{\scriptsize $\pm$0.067}
& 1.371{\scriptsize $\pm$0.056} / 1.745{\scriptsize $\pm$0.080} \\
S-STGCNN 
& 3.092{\scriptsize $\pm$0.309} / 2.325{\scriptsize $\pm$0.301}
& 2.947{\scriptsize $\pm$0.490} / 2.351{\scriptsize $\pm$0.363}
& 2.946{\scriptsize $\pm$0.415} / 2.226{\scriptsize $\pm$0.405}
& 2.946{\scriptsize $\pm$0.378} / 2.188{\scriptsize $\pm$0.278} \\
AgentFormer 
& 0.926{\scriptsize $\pm$0.014} / 0.886{\scriptsize $\pm$0.012}
& 0.919{\scriptsize $\pm$0.008} / 0.869{\scriptsize $\pm$0.006}
& 0.923{\scriptsize $\pm$0.012} / 0.884{\scriptsize $\pm$0.012}
& 0.921{\scriptsize $\pm$0.010} / 0.877{\scriptsize $\pm$0.009} \\
Two-block RNN
& 1.495{\scriptsize $\pm$0.148} / 2.767{\scriptsize $\pm$0.310}
& 1.503{\scriptsize $\pm$0.110} / 2.796{\scriptsize $\pm$0.171}
& 1.516{\scriptsize $\pm$0.134} / 2.805{\scriptsize $\pm$0.256}
& 1.530{\scriptsize $\pm$0.094} / 2.802{\scriptsize $\pm$0.200} \\
Transformer 
& 0.424{\scriptsize $\pm$0.023} / 0.850{\scriptsize $\pm$0.040}
& 0.464{\scriptsize $\pm$0.029} / 0.897{\scriptsize $\pm$0.046}
& 0.331{\scriptsize $\pm$0.036} / 0.645{\scriptsize $\pm$0.091}
& 0.534{\scriptsize $\pm$0.060} / 1.076{\scriptsize $\pm$0.101} \\
S-Implicit 
& 1.067{\scriptsize $\pm$0.149} / 1.790{\scriptsize $\pm$0.383}
& 1.197{\scriptsize $\pm$0.027} / 2.130{\scriptsize $\pm$0.060}
& 1.084{\scriptsize $\pm$0.175} / 1.850{\scriptsize $\pm$0.401}
& 1.016{\scriptsize $\pm$0.224} / 1.776{\scriptsize $\pm$0.451} \\
MSTF
& 0.403{\scriptsize $\pm$0.016} / 0.606{\scriptsize $\pm$0.023}
& 0.406{\scriptsize $\pm$0.018} / 0.603{\scriptsize $\pm$0.013}
& 0.399{\scriptsize $\pm$0.013} / 0.596{\scriptsize $\pm$0.015}
& 0.398{\scriptsize $\pm$0.014} / 0.601{\scriptsize $\pm$0.023} \\
TUTR 
& \underline{0.282}{\scriptsize $\pm$0.015} / \underline{0.477}{\scriptsize $\pm$0.027}
& \underline{0.293}{\scriptsize $\pm$0.008} / \underline{0.481}{\scriptsize $\pm$0.024}
& \underline{0.292}{\scriptsize $\pm$0.013} / \underline{0.496}{\scriptsize $\pm$0.028}
& \underline{0.324}{\scriptsize $\pm$0.019} / \underline{0.518}{\scriptsize $\pm$0.030} \\
\midrule
\textbf{DIDAM-Net} 
& \textbf{0.252}{\scriptsize $\pm$0.022} / \textbf{0.461}{\scriptsize $\pm$0.028}
& \textbf{0.272}{\scriptsize $\pm$0.019} / \textbf{0.465}{\scriptsize $\pm$0.028}
& \textbf{0.265}{\scriptsize $\pm$0.022} / \textbf{0.444}{\scriptsize $\pm$0.028}
& \textbf{0.264}{\scriptsize $\pm$0.015} / \textbf{0.399}{\scriptsize $\pm$0.027} \\
\bottomrule
\end{tabular}
\end{adjustbox}
}
\tabnote{\raggedright\footnotesize $\bullet$ Best in \textbf{bold}; second-best \underline{underlined}.\par}
\label{tab:imputation_main}
\end{table}

Among the baselines, TUTR is the most competitive, ranking second across all imputation settings.
The Transformer exhibits greater performance variation across imputation methods, performing better with Kalman filtering than with Gaussian process imputation. 
AgentFormer achieves intermediate performance, while Two-block RNN is less competitive; both show relatively limited sensitivity to the imputation strategy. 
Graph-based approaches (S-STGCNN and STGAT) yield moderate but relatively stable results, suggesting limited benefits from interaction modeling in this dataset. 
S-Implicit shows noticeable performance variation across imputation methods, whereas MSTF is comparatively insensitive to the imputation strategy but remains less accurate than the leading models. The EKF baseline also yields relatively large prediction errors, indicating limitations of its motion assumptions for these AIS trajectories. 

\begin{figure}[htbp]
  \centering
  \subfloat[Mild imputations.\label{fig:qual_a}]{%
    \includegraphics[width=0.47\linewidth]{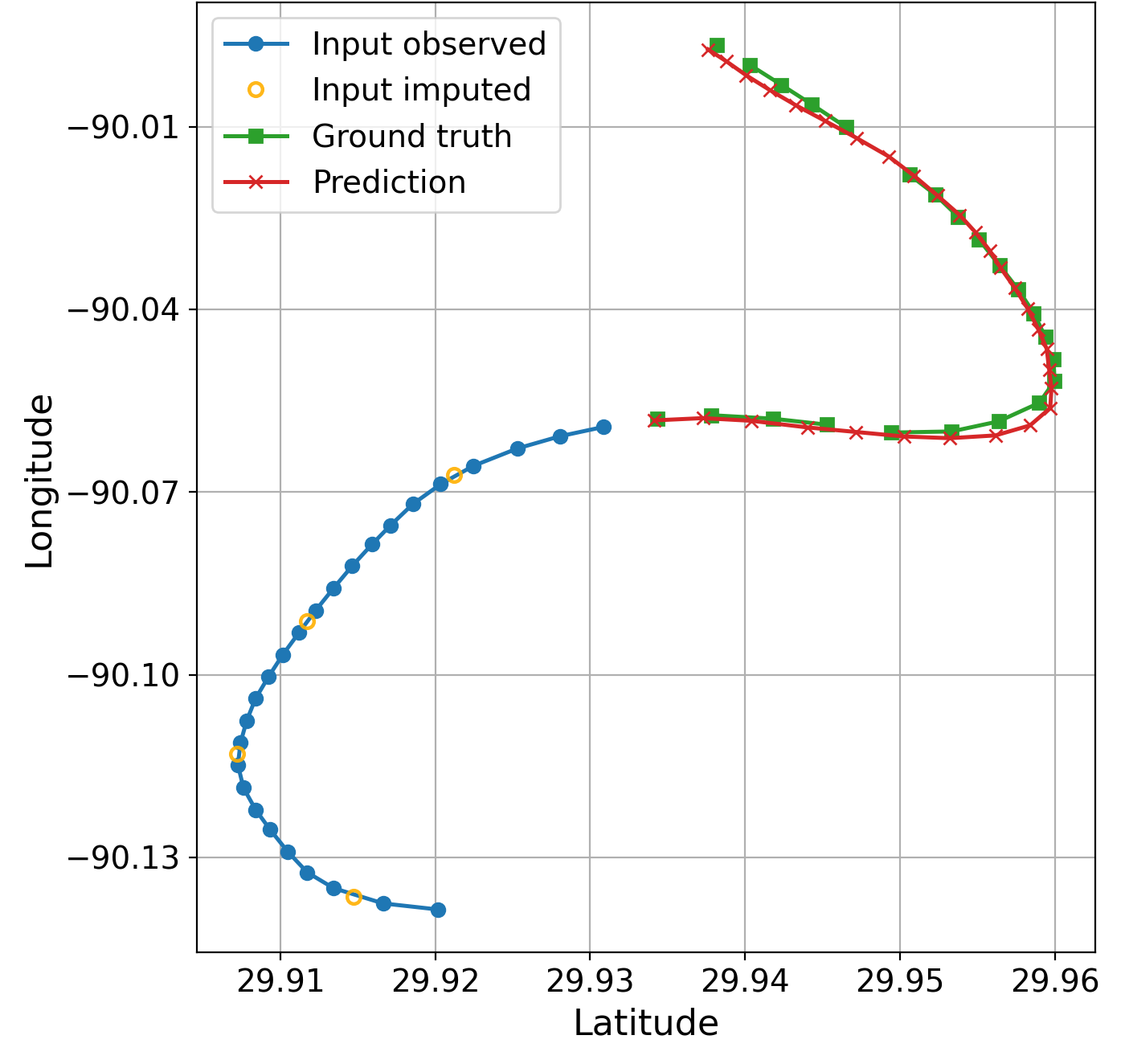}}
  \hfill
  \subfloat[Severe imputations.\label{fig:qual_b}]{%
    \includegraphics[width=0.47\linewidth]{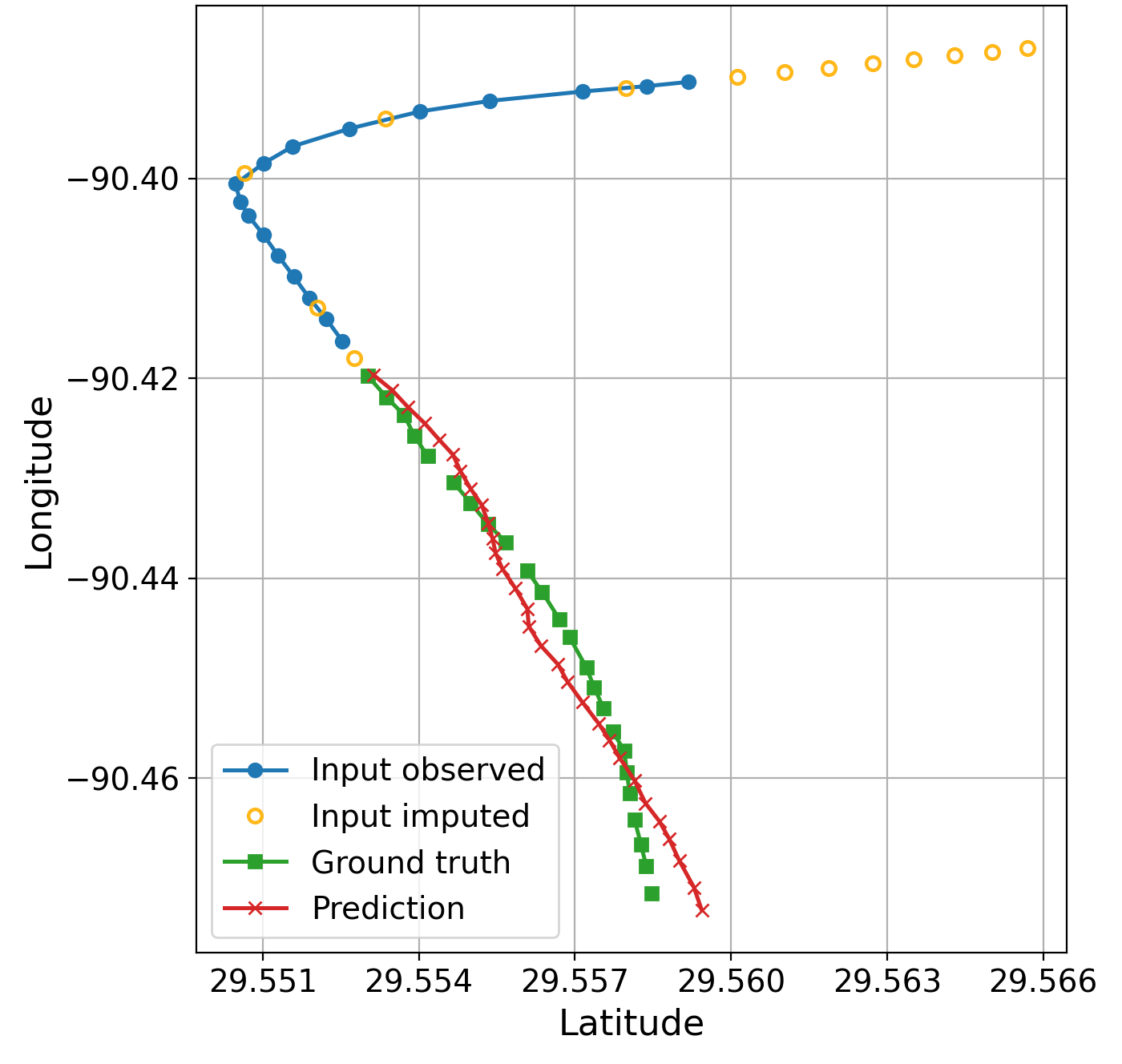}}
  \caption{Qualitative examples of vessel trajectory prediction under mild and severe input imputation.}
  \label{fig:qualitative}
\end{figure}

Figure~\ref{fig:qualitative} illustrates DIDAM-Net's predictions under different proportions of imputed input positions. 
Blue and amber markers indicate observed and imputed input positions, respectively, while red and green trajectories represent the predictions and ground truth. 
Despite differences in input missingness, the predicted trajectories closely follow the ground-truth paths, including curved and turning segments. These examples illustrate how DIDAM-Net can incorporate imputed positions while adaptively discounting their contributions based on the missingness pattern.

\paragraph*{Trajectory prediction under prolonged missing gaps.}
Table~\ref{tab:mingap5_allbaselines} evaluates prediction robustness under prolonged missing gaps. In this stress-test setting, all missing blocks shorter than five timestamps are extended to at least five timestamps on the resampled AIS time grid, resulting in 70.76\% imputed timestamps and substantially reduced availability of observed AIS data. The extended gaps retain their original temporal locations, representing prolonged interruptions in AIS reporting. 
Because this setting uses a separately constructed dataset and train/validation/test split, absolute errors should be compared among methods within each table rather than directly across Tables~\ref{tab:mingap5_allbaselines} and \ref{tab:imputation_main}.

\begin{table}[t]
\centering
\tbl{Trajectory prediction performance under prolonged missing gaps (mean $\pm$ SE of ADE$\downarrow$ / FDE$\downarrow$, in NM).}{%
\setlength{\tabcolsep}{4pt}
\renewcommand{\arraystretch}{1.2}
\begin{adjustbox}{max width=\textwidth}
\begin{tabular}{lcccc}

\toprule
Method & Linear & Cubic & Kalman filter & Gaussian process \\
\midrule
EKF 
& 2.331{\scriptsize $\pm$0.000} / 3.424{\scriptsize $\pm$0.000}
& 2.546{\scriptsize $\pm$0.000} / 3.824{\scriptsize $\pm$0.000}
& 2.246{\scriptsize $\pm$0.000} / 3.178{\scriptsize $\pm$0.000}
& 2.960{\scriptsize $\pm$0.000} / 4.854{\scriptsize $\pm$0.000} \\
STGAT 
& 1.404{\scriptsize $\pm$0.042} / 0.755{\scriptsize $\pm$0.020}
& 1.427{\scriptsize $\pm$0.053} / 0.765{\scriptsize $\pm$0.023}
& 1.469{\scriptsize $\pm$0.084} / 0.789{\scriptsize $\pm$0.043}
& 1.405{\scriptsize $\pm$0.060} / 0.887{\scriptsize $\pm$0.302} \\
S-STGCNN 
& 1.374{\scriptsize $\pm$0.437} / 1.207{\scriptsize $\pm$0.618}
& 1.199{\scriptsize $\pm$0.213} / 0.934{\scriptsize $\pm$0.226}
& 1.139{\scriptsize $\pm$0.064} / 0.900{\scriptsize $\pm$0.128}
& 1.204{\scriptsize $\pm$0.094} / 0.916{\scriptsize $\pm$0.097} \\
AgentFormer 
& 0.912{\scriptsize $\pm$0.009} / 0.887{\scriptsize $\pm$0.013}
& 0.909{\scriptsize $\pm$0.006} / 0.880{\scriptsize $\pm$0.007}
& 0.977{\scriptsize $\pm$0.059} / 0.963{\scriptsize $\pm$0.062}
& 0.981{\scriptsize $\pm$0.058} / 0.964{\scriptsize $\pm$0.062} \\
Two-block RNN
& 1.484{\scriptsize $\pm$0.276} / 2.761{\scriptsize $\pm$0.525}
& 1.456{\scriptsize $\pm$0.118} / 2.681{\scriptsize $\pm$0.227}
& 1.496{\scriptsize $\pm$0.125} / 2.691{\scriptsize $\pm$0.310}
& 1.523{\scriptsize $\pm$0.092} / 2.592{\scriptsize $\pm$0.232} \\
Transformer 
& 0.365{\scriptsize $\pm$0.061} / 0.665{\scriptsize $\pm$0.110}
& 0.611{\scriptsize $\pm$0.276} / 1.137{\scriptsize $\pm$0.506}
& 0.697{\scriptsize $\pm$0.220} / 1.340{\scriptsize $\pm$0.482}
& 0.529{\scriptsize $\pm$0.109} / 0.969{\scriptsize $\pm$0.256} \\
S-Implicit 
& 1.004{\scriptsize $\pm$0.207} / 1.659{\scriptsize $\pm$0.490}
& 0.996{\scriptsize $\pm$0.203} / 1.532{\scriptsize $\pm$0.458}
& 1.087{\scriptsize $\pm$0.273} / 1.929{\scriptsize $\pm$0.514}
& 1.197{\scriptsize $\pm$0.154} / 2.089{\scriptsize $\pm$0.346} \\
MSTF
& 0.666{\scriptsize $\pm$0.020} / 0.901{\scriptsize $\pm$0.027}
& 0.671{\scriptsize $\pm$0.011} / 0.905{\scriptsize $\pm$0.018}
& 0.665{\scriptsize $\pm$0.010} / 0.907{\scriptsize $\pm$0.017}
& 0.667{\scriptsize $\pm$0.016} / 0.908{\scriptsize $\pm$0.025} \\
TUTR 
& \underline{0.268}{\scriptsize $\pm$0.010} / \underline{0.415}{\scriptsize $\pm$0.022}
& \underline{0.332}{\scriptsize $\pm$0.006} / \underline{0.505}{\scriptsize $\pm$0.015}
& \underline{0.277}{\scriptsize $\pm$0.015} / \underline{0.441}{\scriptsize $\pm$0.034}
& \underline{0.325}{\scriptsize $\pm$0.021} / \underline{0.486}{\scriptsize $\pm$0.039} \\
\midrule
\textbf{DIDAM-Net} 
& \textbf{0.226}{\scriptsize $\pm$0.012} / \textbf{0.381}{\scriptsize $\pm$0.028}
& \textbf{0.262}{\scriptsize $\pm$0.020} / \textbf{0.379}{\scriptsize $\pm$0.029}
& \textbf{0.199}{\scriptsize $\pm$0.020} / \textbf{0.295}{\scriptsize $\pm$0.032}
& \textbf{0.262}{\scriptsize $\pm$0.020} / \textbf{0.374}{\scriptsize $\pm$0.024} \\
\bottomrule
\end{tabular}
\end{adjustbox}
}
\tabnote{\raggedright\footnotesize $\bullet$ Best in \textbf{bold}; second-best \underline{underlined}.\par}
\label{tab:mingap5_allbaselines}
\end{table}

DIDAM-Net maintains the lowest errors across all imputation methods, maintaining a consistent advantage over the baselines despite the reduced availability of observed positions. These results suggest that dynamic attention discounting is particularly beneficial when trajectory histories contain prolonged missing intervals.

Among the baselines, TUTR consistently ranks second across all imputation methods.
The Transformer shows higher errors and greater sensitivity to the imputation strategy. 
STGAT attains competitive FDE but substantially higher ADE, indicating less consistent accuracy across the prediction horizon.
Two-block RNN and MSTF show limited sensitivity to the imputation method but remain less accurate than the leading models.
S-Implicit exhibits substantial performance variation across imputations, while S-STGCNN yields relatively high prediction errors. 
The EKF baseline performs least competitively, suggesting limitations of its motion assumptions under prolonged missing gaps.

Overall, these results demonstrate DIDAM-Net's robustness to prolonged missing intervals and different imputation strategies. Its consistent performance supports the effectiveness of selectively discounting imputed positions, rather than treating observed and imputed inputs equally, when genuine AIS observations are sparse.

\subsection{Ablation study}
This section examines the contributions of DIDAM-Net's components and its performance under different prediction settings. The ablation studies investigate (1) the effectiveness of attention masking at different locations in the architecture, (2) the contributions of individual imputation metadata features to mask generation, and (3) the effect of prediction horizon on accuracy.
All experiments use cubic imputation for consistency, with performance measured using the best-of-20 ADE/FDE metrics defined in Section~\ref{sec:implementation-details}, evaluated only at observed timestamps.

\paragraph*{Impact of attention masking components.}
The first ablation examines the contribution of DIDAM at different stages of the architecture. Table~\ref{tab:ablation_mask} compares four configurations: (i) \textit{No Mask}, without DIDAM, (ii) \textit{Encoder Only}, with DIDAM applied to encoder self-attention, (iii) \textit{Decoder Only}, with DIDAM applied to decoder cross-attention, and (iv) \textit{Full Mask}, with DIDAM applied to both encoder and decoder as proposed. 
Results are reported for DIDAM-Net, Transformer, and AgentFormer, each of which supports attention masking at the encoder and decoder stages, enabling a consistent comparison of masking configurations.

\begin{table}[t]
\centering
\tbl{Impact of attention masking components on trajectory prediction accuracy (mean $\pm$ SE of ADE$\downarrow$ / FDE$\downarrow$, in NM).}{%
\setlength{\tabcolsep}{4pt}      
\renewcommand{\arraystretch}{1.2}
\begin{adjustbox}{max width=\textwidth}

\begin{tabular}{lccc}
\toprule
Configuration & DIDAM-Net & Transformer & AgentFormer \\
\midrule
No Mask      
& \underline{0.301}{\scriptsize $\pm$0.020} / \underline{0.505}{\scriptsize $\pm$0.032}
& 0.464{\scriptsize $\pm$0.029} / 0.897{\scriptsize $\pm$0.046}
& 0.918{\scriptsize $\pm$0.008} / \textbf{0.871}{\scriptsize $\pm$0.007} \\
Encoder Only 
& 0.317{\scriptsize $\pm$0.021} / 0.529{\scriptsize $\pm$0.033}
& \underline{0.427}{\scriptsize $\pm$0.008} / 0.851{\scriptsize $\pm$0.047}
& 0.914{\scriptsize $\pm$0.012} / \underline{0.873}{\scriptsize $\pm$0.009} \\
Decoder Only 
& 0.321{\scriptsize $\pm$0.025} / 0.532{\scriptsize $\pm$0.038}
& 0.429{\scriptsize $\pm$0.036} / \underline{0.849}{\scriptsize $\pm$0.086}
& \underline{0.912}{\scriptsize $\pm$0.009} / 0.874{\scriptsize $\pm$0.006} \\
Full Mask    
& \textbf{0.272}{\scriptsize $\pm$0.019} / \textbf{0.465}{\scriptsize $\pm$0.028}
& \textbf{0.398}{\scriptsize $\pm$0.031} / \textbf{0.777}{\scriptsize $\pm$0.080}
& \textbf{0.909}{\scriptsize $\pm$0.006} / 0.875{\scriptsize $\pm$0.009} \\
\bottomrule
\end{tabular}
\end{adjustbox}
}
\tabnote{\raggedright\footnotesize $\bullet$ Best in \textbf{bold}; second-best \underline{underlined}.\par}
\label{tab:ablation_mask}
\end{table}

The results reveal differences in the effectiveness of attention masking across architectures. 
For DIDAM-Net, applying masking to either encoder self-attention or decoder cross-attention alone slightly increases prediction errors relative to the no-mask configuration. 
In contrast, the \emph{Full Mask} configuration achieves the lowest errors, suggesting a complementary effect between encoder and decoder masking. Encoder masking adjusts the representations learned from imputed trajectories, while decoder masking discounts potentially unreliable encoder positions during cross-attention.

The Transformer exhibits a slightly different pattern: encoder-only and decoder-only masking each yield modest improvements over the no-mask configuration, while combining both produces the largest gains. 
AgentFormer, however, shows only marginal changes, with  slight improvements in ADE and nearly unchanged FDE across configurations.

Overall, these results suggest that the effectiveness of DIDAM depends on the underlying architecture and the placement of attention masking. 
For DIDAM-Net, the greatest benefit arises from combining encoder and decoder masking, highlighting the importance of their joint application rather than either component in isolation.

\paragraph*{Importance of imputation metadata features.}
Building on the dual-masking ablation, the next ablation investigates the contributions of the three imputation metadata feature groups introduced in Section~\ref{sec:methodology}: block structure, temporal distance, and context. 
Two complementary settings are considered: (i) \emph{leave-one-group-out}, in which one feature group is removed while the other two are retained, and (ii) \emph{single-group-only}, in which only one feature group is retained. Table~\ref{tab:ablation_metadata} reports the results.

\begin{table}[t]
\centering
\tbl{Ablation study of imputation metadata features\\ (mean $\pm$ SE of ADE$\downarrow$ / FDE$\downarrow$, in NM).}{%
\setlength{\tabcolsep}{6pt}
\renewcommand{\arraystretch}{1.2}
\begin{tabular}{lc}
\toprule
Configuration & ADE / FDE \\
\midrule
\multicolumn{2}{l}{\textit{Full Model}} \\
DIDAM-Net (B+T+C) 
& \textbf{0.272}{\scriptsize $\pm$0.019} / \textbf{0.465}{\scriptsize $\pm$0.028} \\
\midrule
\multicolumn{2}{l}{\textit{Leave-one-group-out}} \\
Without Block    
& 0.325{\scriptsize $\pm$0.018} / 0.535{\scriptsize $\pm$0.044} \\
Without Temporal 
& 0.327{\scriptsize $\pm$0.022} / 0.530{\scriptsize $\pm$0.044} \\
Without Context  
& 0.307{\scriptsize $\pm$0.021} / 0.512{\scriptsize $\pm$0.046} \\
\midrule
\multicolumn{2}{l}{\textit{Single-group-only}} \\
Block only       
& \underline{0.294}{\scriptsize $\pm$0.023} / \underline{0.486}{\scriptsize $\pm$0.038} \\
Temporal only    
& 0.323{\scriptsize $\pm$0.030} / 0.528{\scriptsize $\pm$0.042} \\
Context only     
& 0.332{\scriptsize $\pm$0.024} / 0.560{\scriptsize $\pm$0.047} \\
\bottomrule
\end{tabular}
}
\tabnote{\footnotesize B: Block structure, T: Temporal distance, C: Context.}
\tabnote{\raggedright\footnotesize $\bullet$ Best in \textbf{bold}; second-best \underline{underlined}.\par}
\label{tab:ablation_metadata}
\end{table}

In the leave-one-group-out setting, removing either block structure  or temporal distance produces the largest performance degradation, with comparable effects for both groups. 
Removing context features also increases prediction errors, but to a lesser extent.
These results suggest that block structure and temporal distance provide particularly important information for generating effective attention masks, while context offers additional benefits.

Among the single-group configurations, \emph{Block only} achieves the lowest errors, followed by \emph{Temporal only}, whereas \emph{Context only} performs least competitively. This indicates that missing-block structure provides the strongest individual contribution, with temporal distance and context offering complementary information when combined with other features.

The full model achieves the lowest ADE and FDE across all configurations. Although its improvements over the strongest subsets are modest, the results support the value of combining all three feature groups: block structure characterizes missing intervals, temporal distance captures proximity to observed positions, and context describes the surrounding observation patterns. Together, these features enable more effective attention discounting under missing AIS observations.

\paragraph*{Performance across different prediction horizons.}
The final ablation examines DIDAM-Net's performance across prediction horizons of 10, 20, and 30 timestamps, using a fixed observation window of 30 timestamps. As shown in Table~\ref{tab:ablation_horizon}, DIDAM-Net consistently achieves the lowest ADE and FDE across all horizons, with its advantage over the baselines becoming more pronounced as the prediction horizon increases. At 10 timestamps, DIDAM-Net slightly outperforms TUTR, while the performance gap widens at 20 and 30 timestamps as the baselines exhibit greater increases in prediction errors.

\begin{table}[t]
\centering
\tbl{Trajectory prediction performance across different prediction horizons (mean $\pm$ SE of ADE$\downarrow$ / FDE$\downarrow$, in NM).}{%
\setlength{\tabcolsep}{3pt}      
\renewcommand{\arraystretch}{1.2}
\begin{adjustbox}{max width=\textwidth}
\begin{tabular}{lcccccc}
\toprule
\multirow{2}{*}{Method} & \multicolumn{2}{c}{10 timestamps} & \multicolumn{2}{c}{20 timestamps} & \multicolumn{2}{c}{30 timestamps} \\
\cmidrule(lr){2-3} \cmidrule(lr){4-5} \cmidrule(lr){6-7}
 & ADE & FDE & ADE & FDE & ADE & FDE \\
\midrule
EKF 
& 1.722{\scriptsize $\pm$0.000} & 1.999{\scriptsize $\pm$0.000}
& 2.258{\scriptsize $\pm$0.000} & 3.662{\scriptsize $\pm$0.000}
& 3.221{\scriptsize $\pm$0.000} & 6.345{\scriptsize $\pm$0.000} \\
STGAT 
& 0.410{\scriptsize $\pm$0.105} & 0.525{\scriptsize $\pm$0.116}
& 0.956{\scriptsize $\pm$0.035} & 1.233{\scriptsize $\pm$0.030}
& 1.360{\scriptsize $\pm$0.047} & 1.714{\scriptsize $\pm$0.056} \\
S-STGCNN 
& 1.423{\scriptsize $\pm$0.121} & 1.061{\scriptsize $\pm$0.106}
& 1.985{\scriptsize $\pm$0.132} & 1.572{\scriptsize $\pm$0.162}
& 2.947{\scriptsize $\pm$0.490} & 2.351{\scriptsize $\pm$0.363} \\
AgentFormer 
& 0.932{\scriptsize $\pm$0.016} & 0.948{\scriptsize $\pm$0.020}
& 0.912{\scriptsize $\pm$0.019} & 0.898{\scriptsize $\pm$0.021}
& 0.919{\scriptsize $\pm$0.008} & 0.869{\scriptsize $\pm$0.006} \\
Two-block RNN
& 0.501{\scriptsize $\pm$0.029} & 0.907{\scriptsize $\pm$0.056}
& 1.012{\scriptsize $\pm$0.060} & 1.868{\scriptsize $\pm$0.118}
& 1.503{\scriptsize $\pm$0.110} & 2.796{\scriptsize $\pm$0.171} \\
Transformer
& 0.097{\scriptsize $\pm$0.006} & 0.165{\scriptsize $\pm$0.013}
& 0.296{\scriptsize $\pm$0.092} & 0.587{\scriptsize $\pm$0.209}
& 0.464{\scriptsize $\pm$0.029} & 0.897{\scriptsize $\pm$0.046} \\
S-Implicit 
& 0.479{\scriptsize $\pm$0.077} & 0.833{\scriptsize $\pm$0.140}
& 0.898{\scriptsize $\pm$0.036} & 1.616{\scriptsize $\pm$0.065}
& 1.197{\scriptsize $\pm$0.027} & 2.130{\scriptsize $\pm$0.060} \\
MSTF
& 0.127{\scriptsize $\pm$0.002} & 0.172{\scriptsize $\pm$0.005}
& 0.264{\scriptsize $\pm$0.008} & 0.388{\scriptsize $\pm$0.013}
& 0.406{\scriptsize $\pm$0.018} & 0.603{\scriptsize $\pm$0.013} \\
TUTR 
& \underline{0.093}{\scriptsize $\pm$0.002} & \underline{0.147}{\scriptsize $\pm$0.005}
& \underline{0.184}{\scriptsize $\pm$0.008} & \underline{0.299}{\scriptsize $\pm$0.015}
& \underline{0.293}{\scriptsize $\pm$0.008} & \underline{0.481}{\scriptsize $\pm$0.024} \\
\midrule
\textbf{DIDAM-Net} 
& \textbf{0.092}{\scriptsize $\pm$0.003} & \textbf{0.139}{\scriptsize $\pm$0.006}
& \textbf{0.175}{\scriptsize $\pm$0.007} & \textbf{0.285}{\scriptsize $\pm$0.012}
& \textbf{0.272}{\scriptsize $\pm$0.019} & \textbf{0.465}{\scriptsize $\pm$0.028} \\
\bottomrule
\end{tabular}
\end{adjustbox}
}
\tabnote{\raggedright\footnotesize $\bullet$ Best in \textbf{bold}; second-best \underline{underlined}.\par}
\label{tab:ablation_horizon}
\end{table}

Among the baselines, TUTR remains the strongest competitor, ranking second at 20 and 30 timestamps. The Transformer exhibits more pronounced performance degradation as the horizon increases. Graph-based methods are less competitive, with S-STGCNN showing substantial error increases and STGAT exhibiting relatively stable but higher errors. MSTF maintains comparatively strong performance, although its errors increase more rapidly than those of DIDAM-Net. Two-block RNN achieves intermediate performance, while EKF exhibits the greatest deterioration at longer horizons.

Overall, these results demonstrate DIDAM-Net's effectiveness across prediction horizons, particularly for longer-term forecasting. Its increasing advantage over the baselines is consistent with the potential benefit of adaptive attention discounting in limiting the influence of imputation errors on successive predictions. This capability may be useful for downstream applications requiring multi-horizon forecasts, such as collision-risk assessment and route planning. However, these applications are beyond the scope of this study, and collision avoidance would additionally require modeling vessel--vessel interactions, which are not considered in the present single-vessel formulation.

\subsection{Inference time}
\label{sec:cost}
The computational feasibility of DIDAM-Net for real-time trajectory prediction was assessed by measuring inference time using the model trained with cubic imputation and a 30-timestamp prediction horizon (Table~\ref{tab:imputation_main}). 
Test windows sharing the same final observation timestamp were processed as a single batch, representing vessels requiring simultaneous predictions at each AIS update. 
Across 699 test updates, an average of 26.5 vessels (maximum 49) were processed simultaneously, generating one predicted trajectory per vessel.
Median inference latency was 201~ms, and the maximum was 0.31~s, both accounting for less than 1\% of the one-minute AIS update interval.
Compared with the unmasked configuration in  Table~\ref{tab:ablation_mask}, DIDAM increased inference time by approximately 7\%, indicating modest computational overhead. These results suggest that DIDAM-Net can support trajectory prediction within a one-minute update cycle under the evaluated conditions.

\section{Discussion}
The experimental results demonstrate that DIDAM-Net consistently outperforms existing trajectory prediction methods across imputation strategies and missingness conditions, with particularly pronounced advantages under prolonged missing gaps. Its relatively stable performance across imputation methods suggests that dynamic attention discounting reduces sensitivity to the choice of imputation strategy, allowing the model to make effective use of imputed trajectories without treating all input positions equally. This finding is also relevant to uncertainty-aware prediction, where accounting for differences in observation reliability is important for robust inference. Although DIDAM-Net does not explicitly quantify predictive uncertainty \citep{pmlr-v235-yoon24a,yoon2026courtroom,yoon2026uncertainty}, its attention-discounting mechanism incorporates information related to the potential unreliability of imputed observations. 

DIDAM-Net's advantage is especially evident when consecutive missing gaps are prolonged and at longer prediction horizons. These findings suggest that selectively discounting potentially unreliable imputed positions helps mitigate their influence on trajectory predictions when observed information is limited. By combining Transformer-based temporal representations with LSTM-based sequential prediction, DIDAM-Net leverages historical trajectory information while adaptively adjusting attention according to imputation metadata. This design provides an alternative to missing-aware approaches that exclude imputed values and imputation-aware approaches that do not explicitly account for differences in their reliability.

The ablation studies further clarify the contributions of DIDAM's components. Applying attention masking jointly in the encoder and decoder yields the largest improvements, suggesting complementary roles: encoder masking adjusts temporal representations based on missingness patterns, while decoder masking discounts potentially unreliable encoder positions during cross-attention. The metadata ablation identifies missing-block structure as the strongest individual feature group, with temporal distance and context providing complementary information. Together, these results highlight the importance of both masking placement and missingness-aware feature design.

From an operational perspective, DIDAM-Net offers several practical advantages. Its consistent performance across imputation strategies suggests flexibility in selecting imputation methods according to application requirements. Furthermore, the inference-time results (Section~\ref{sec:cost}) indicate that predictions for multiple vessels can be generated within a one-minute AIS update interval under the evaluated conditions, with modest overhead from DIDAM. These properties support its potential integration into maritime trajectory-monitoring systems.

Despite these advantages, several limitations remain. First, DIDAM-Net processes vessels independently and does not explicitly model vessel--vessel interactions, which may be important in congested waterways. Second, the evaluation is restricted to a single geographical region, where towing vessels constitute a substantial proportion of the vessel population (Table~\ref{tab:population}). Generalization to other maritime environments, missingness patterns, and vessel types, such as open-sea container ships and high-speed passenger vessels, as well as underrepresented patterns and rare navigation events \citep{park2022prediction, cho2023prediction}, therefore remains to be established. 
Extending DIDAM-Net to incorporate vessel interactions and evaluating its performance across more diverse maritime settings are promising directions for future research.

\section{Conclusion}
This paper introduced DIDAM-Net, a vessel trajectory prediction model designed to address long consecutive missing gaps in AIS data through a Dynamic Imputation-Discounting Attention Mask. Unlike conventional imputation-aware methods that do not explicitly distinguish among imputed values based on their missingness patterns, DIDAM-Net adaptively discounts attention using imputation metadata. This approach retains the temporal continuity provided by imputation while reducing the influence of potentially unreliable imputed observations.

Comprehensive experiments on real-world AIS data from the Lower Mississippi River Delta demonstrate DIDAM-Net's effectiveness across diverse imputation strategies and missing patterns, with particularly pronounced advantages under prolonged missing gaps and longer prediction horizons. 
Ablation studies further highlight the complementary contributions of encoder and decoder masking and the importance of missing-block structure and temporal metadata in generating attention discounts. These findings suggest that incorporating missingness patterns into attention mechanisms is a promising approach to trajectory prediction under incomplete observations. The resulting improvements may also benefit downstream applications such as collision-risk assessment and route planning, although these applications were not evaluated and would require additional capabilities, including vessel--vessel interaction modeling. 

Future research directions include evaluating generalization across diverse geographical regions and vessel populations, incorporating vessel-specific characteristics and interactions, and developing data-driven alternatives to hand-crafted imputation metadata features. Extending the proposed approach to other irregularly observed time-series domains, such as clinical monitoring, satellite observations, and IoT sensor networks, also presents a promising direction for further investigation.

\section*{Author contributions}
\noindent\textbf{Sanha Chang:} Conceptualization, Methodology, Software, Formal analysis, Writing -- original draft.\par
\noindent\textbf{Heeyoung Kim:} Supervision, Funding acquisition, Writing -- review \& editing.

\section*{Disclosure statement}
The authors declare that they have no known competing financial interests or personal relationships that could have appeared to influence the work reported in this paper.

\section*{Data availability statement}
The AIS vessel traffic data (NAIS 2023) used in this study are publicly available from the NOAA Office for Coastal Management via the MarineCadastre.gov data portal \citep{noaa_nais_2023}.

\section*{Funding}
This work was supported by the National Research Foundation of Korea (NRF) grant funded by the Korea government (MSIT) (2023R1A2C2005453, RS-2023-00218913).

\bibliographystyle{tfcse}
\bibliography{references}

\end{document}